\documentclass[preprint]{elsarticle}

\usepackage{graphicx}

\usepackage{caption}
\usepackage{setspace}
\usepackage{epsfig}
\usepackage{graphicx}
\usepackage{subfigure}
\usepackage{subfloat}
\usepackage{amsmath}
\usepackage{amssymb}
\usepackage{booktabs} 
\usepackage{multirow}
\usepackage[ruled]{algorithm2e}
\usepackage{bm}
\usepackage{threeparttable}
\newcommand{\etal}{\textit{et al. }}

\biboptions{sort&compress}

\begin{document}
\captionsetup[figure]{labelfont={bf},labelsep=period,name={Fig.}}
\captionsetup[table]{labelfont={bf},labelsep=newline,justification=justified,singlelinecheck= false}
\begin{frontmatter}

\title{PoseConvGRU: A Monocular Approach for Visual Ego-motion Estimation by Learning}

\author{Guangyao Zhai\tnoteref{myfootnote}}
\author{Liang Liu\tnoteref{myfootnote}}
\author{Linjian Zhang}
\author{Yong Liu\corref{mycorrespondingauthor}}
\ead{yongliu@iipc.zju.edu.cn}
\cortext[mycorrespondingauthor]{Corresponding author}
\tnotetext[myfootnote]{Guangyao Zhai and Liang Liu contribute equally to this work.}
\address{Institute of Cyber-Systems
and Control, Zhejiang University}

\begin{abstract}
While many visual ego-motion algorithm variants have been proposed in the past decade, learning based ego-motion estimation methods have seen an increasing attention because of its desirable properties of robustness to image noise and camera calibration independence. In this work, we propose a data-driven approach of fully trainable visual ego-motion estimation for a monocular camera. We use an end-to-end learning approach in allowing the model to map directly from input image pairs to an estimate of ego-motion (parameterized as 6-DoF transformation matrices). We introduce a novel two-module Long-term Recurrent Convolutional Neural Networks called PoseConvGRU, with an explicit sequence pose estimation loss to achieve this. The feature-encoding module encodes the short-term motion feature in an image pair, while the memory-propagating module captures the long-term motion feature in the consecutive image pairs. The visual memory is implemented with convolutional gated recurrent units, which allows propagating information over time. At each time step, two consecutive RGB images are stacked together to form a 6 channels tensor for module-1 to learn how to extract motion information and estimate poses. The sequence of output maps is then passed through a stacked ConvGRU module to generate the relative transformation pose of each image pair. We also augment the training data by randomly skipping frames to simulate the velocity variation which results in a better performance in turning and high-velocity situations. Randomly horizontal flipping and temporal flipping of the sequences is also performed. We evaluate the performance of our proposed approach on the KITTI Visual Odometry benchmark. The experiments show a competitive performance of the proposed method to the geometric method and encourage further exploration of learning based methods for the purpose of estimating camera ego-motion even though geometrical methods demonstrate promising results.
\end{abstract}

\begin{keyword}
\texttt Ego-motion\sep  Pose estimation\sep Deep learning\sep  Convolutional Neural Networks\sep Recurrent Convolutional Networks
\end{keyword}

\end{frontmatter}



\section{Introduction}

For autonomous navigation of intelligent vehicles, the ability of vehicle self-localization during its movement is very important. The method, estimating the position of the vehicle by integrating data of various sensors, is called odometry. With the development of computer vision technology, more and more visual sensors are used for vehicle positioning and motion estimation. We refer to the studying problem of obtaining camera pose through vision as VO (Visual Odometry) \cite{nister2004visual} or visual ego-motion \cite{chen2001three}. The visual sensor not only provides rich sensory information, but also has the advantages of low cost and small size. The mainstream visual ego-motion methods mainly estimate camera poses based on the geometrical characteristics of the objects in the images, so the images must contain a large number of stable texture features. Once there is an obstruction in the scene or in a foggy day, and if there are no other sensors (IMU, laser, etc.), the accuracy of the geometric methods is subject to severe interference. Since many other sensors may not be useful in many practical applications, localization and ego-motion estimation technique based on vision methods still has a lot of space for research.

Recently, more and more researchers have paid much attention to the deep learning study \cite{gu2018recent}, \cite{yao2019review}, \cite{erfani2016high}. Developed to present days, kinds of deep learning approaches represented by convolutional neural networks play very important roles in the field of computer vision \cite{yang2019asymmetric}, \cite{bu2019deep}. These deep neural networks are more effective in extracting image features and finding potential patterns than traditional methods. Therefore, some related researchers consider applying deep learning in the field of visual ego-motion research, letting the deep neural network directly learn the geometric relationship through images to realize the end-to-end pose estimation. This end-to-end process completely eliminates the steps of feature extraction, feature matching, camera calibration, and graph optimization in the traditional methods, and directly obtains the camera poses according to the input images.

This paper mainly studies the problem of camera relative pose estimation by deep learning, only considering the situation of monocular VO. We introduce a novel Long-term Recurrent Convolutional Neural Networks, containing two modules, called PoseConvGRU. The feature-encoding module extracts the short-term motion feature in an image pair, while the memory-propagating module captures the long-term motion feature in the consecutive image pairs to estimate camera poses. We sum the l2-loss of 6-DoF pose for each image pair with another loss term on the sequence of adjacent image pairs by compounding the poses estimated from each of the image pair measurements, that mimics the local bundle adjustment optimization in geometric visual odometry to improve the accuracy of the estimation of camera poses and preserve temporal consistency. We take the sequences 00, 01, 02, 08, 09 for training and the 03, 04, 05, 06, 07, 10 for testing, as common practice. The main contributions are as follows:

\begin{itemize}
\item We propose a novel framework named as PoseConvGRU, an approach of visual ego-motion estimation which is data-driven and fully trainable, with an explicit sequence pose estimation loss to mimic the bundle optimization.

\item Our proposed neural network does not matter with the optical flow or other flow-like subspace, unlike other learning-based ego-motion estimation algorithms, which need to spend plenty of time to calculate the pre-processed dense optical flow before training the neural network \cite{costante2015exploring} or use a pre-trained network to estimate the optical flow with additional calculation costs\cite{muller2017flowdometry}, \cite{costante2018ls}.

\item We augment the training data, performed on the KITTI Visual Odometry \cite{Geiger2012CVPR}, by randomly skipping frames to simulate the velocity variation which results in a better performance in turning and high-velocity situations. Randomly horizontal flipping and temporal flipping of the sequences is also performed.

\end{itemize}

\section{Related Work}\label{Related Works}
\subsection{Progress in geometric research}
Matthies \etal proposed to implement robotic indoor navigation through visual input. The main research at that time was based on the NASA Mars Exploration Program \cite{moravec1980obstacle}. The real foundation for the VO problem is a real-time visual odometry designed by Nister \etal \cite{nister2004visual}, which builds its implementary framework.

Based on this framework, the solution of the VO problem can be further divided into two sorts of methods: feature-point methods and direct methods:

\smallskip
\noindent
\textbf{Feature-point methods} mainly extract the feature points in adjacent frame images, calculating the geometric relationship of the feature points by using multi-view geometry to estimate the relative camera poses, such as LIBVISO2 \cite{geiger2011stereoscan}, ORB-SLAM \cite{mur2015orb}. However, these kinds of methods are very time-consuming when extracting features, and we are only concerned about the extracted feature points with the abundant information of other pixels in images ignored, so that features extracted from the images are not sufficient to restore visual ego-motion if the image texture information captured by the camera is scarce, and typically these methods will not work properly.

\smallskip
\noindent
\textbf{Direct methods} work as long as there is a change in the scene. The obvious difference between the direct methods and the feature-point methods is that it is not necessary to calculate the descriptors and the key points, but the visual ego-motion is estimated directly based on the luminance information of the pixels in the images. These methods avoid the prolonged calculation time of features and the lack of features. According to the number of pixels used, the direct methods can be further divided into three types: sparse ones, dense ones, and semi-dense ones. Open-sourced projects using direct methods such as SVO \cite{forster2014svo}, LSD-SLAM \cite{engel2014lsd}, DSO \cite{engel2017direct} have gradually become important parts of the visual ego-motion algorithm.

However, all these geometric research works are very cumbersome and complex, and each module needs fine adjustment to achieve good results in a specific environment. Moreover, the existing frameworks have been basically fixed, the algorithms have almost reached the bottleneck, and the upside space is getting smaller and smaller. To break through this development bottleneck, it still needs to face great challenges.

\subsection{Progress in deep learning}
Roberts \etal \cite{roberts2008memory} proposed to study visual ego-motion problems by a learning method firstly, but this method did not achieve position estimation of the camera 3-DoF poses, and the error was very large. The first use of deep learning to study visual ego-motion problems is a method of using convolutional neural networks to learn visual odometry introduced by Kishore \etal \cite{konda2015learning}, subtly transforming the pose regression problems into classification problems, but the reliability of the obtained result is not high since a large error has been generated in the process of discretizing the direction angle and velocity. The first method for end-to-end estimation of camera 6-dimensional pose based on convolutional neural networks is PoseNet proposed by Kendall \etal \cite{kendall2015posenet}. The neural network framework of this method was modified from GoogLeNet \cite{szegedy2015going}. Due to PoseNet's inaccurate estimation for handling scenes with some obstacles, Kendall \etal \cite{kendall2016modelling} proposed a Bayesian convolutional neural network to regress the camera 6-DoF pose. The advantage of using a Bayesian convolutional neural network is that it can measure the uncertainties of camera's poses and use these uncertainties to estimate the localization error and determine whether the test images is repeated. Mohanty \etal proposed DeepVO \cite{mohanty2016deepvo}. The CNN part of this method is based on AlexNet \cite{krizhevsky2012imagenet}. It inputs two adjacent RGB images and directly estimates the relative pose between the two images in an end-to-end way. For scenes that have not appeared in the training set before, the estimation results are very unsatisfactory. This work attempts to regard the FAST features of the images as additional input information, but it cannot fundamentally solve the scenario migration problem. Costante \etal \cite{costante2015exploring} proposed two CNN structures to estimate the frame-to-frame poses. Since this method requires input of optical flow images, pre-processing of optical flow will cost more calculation resources.

Ronald \etal proposed VINet \cite{clark2017vinet}. It not only integrates IMU information into the deep neural network to study visual ego-motion problems, but also applies sequence learning to consider the pose relationship among multiple frames of images. This paper provides a novel approach to VIO (Visual Inertial Odometry) field, and the combination of CNN and LSTM for sequence learning has contributed greatly to subsequent research. The authors above also propose VidLoc \cite{clark2017vidloc}, using CNN and LSTM to estimate the global poses of consecutive frames, which achieves a significant improvement over PoseNet \cite{kendall2015posenet}. In the same year, another author of the paper published a new version of DeepVO \cite{wang2017deepvo}. The paper uses the image sequences as input. Firstly, the image features are extracted by CNN, puted into the RNN to learn the geometry relationship among successive frames, and then the relative poses of multi-frame are directly output. The relative transformation poses between the images are very positive compared to all the previous research work.

The existing visual ego-motion estimation methods based on convolutional neural network are far less effective than the geometry-based methods. On the basis of utilizing RGB images, there is a kind of method of using optical flow to help obtain ego-motion, like \cite{costante2015exploring} , but the calculation of pre-processed dense optical flow is time consuming, and the accuracy of optical flow calculation has a great influence on the visual ego-motion estimation, so this sort of method is difficult to be widely used. Our visual ego-motion estimation method based on recurrent convolutional neural network \cite{mohanty2016deepvo} performs better than monocular VISO2, but there is still a gap compared with stereo version. The selection of image sequences and the design of loss function all have a great influence on the experimental results in this method, so there is still a lot of space for improvement on this foundation.

\begin{figure}[ht]
    \centering
    \includegraphics[width=4.5in]{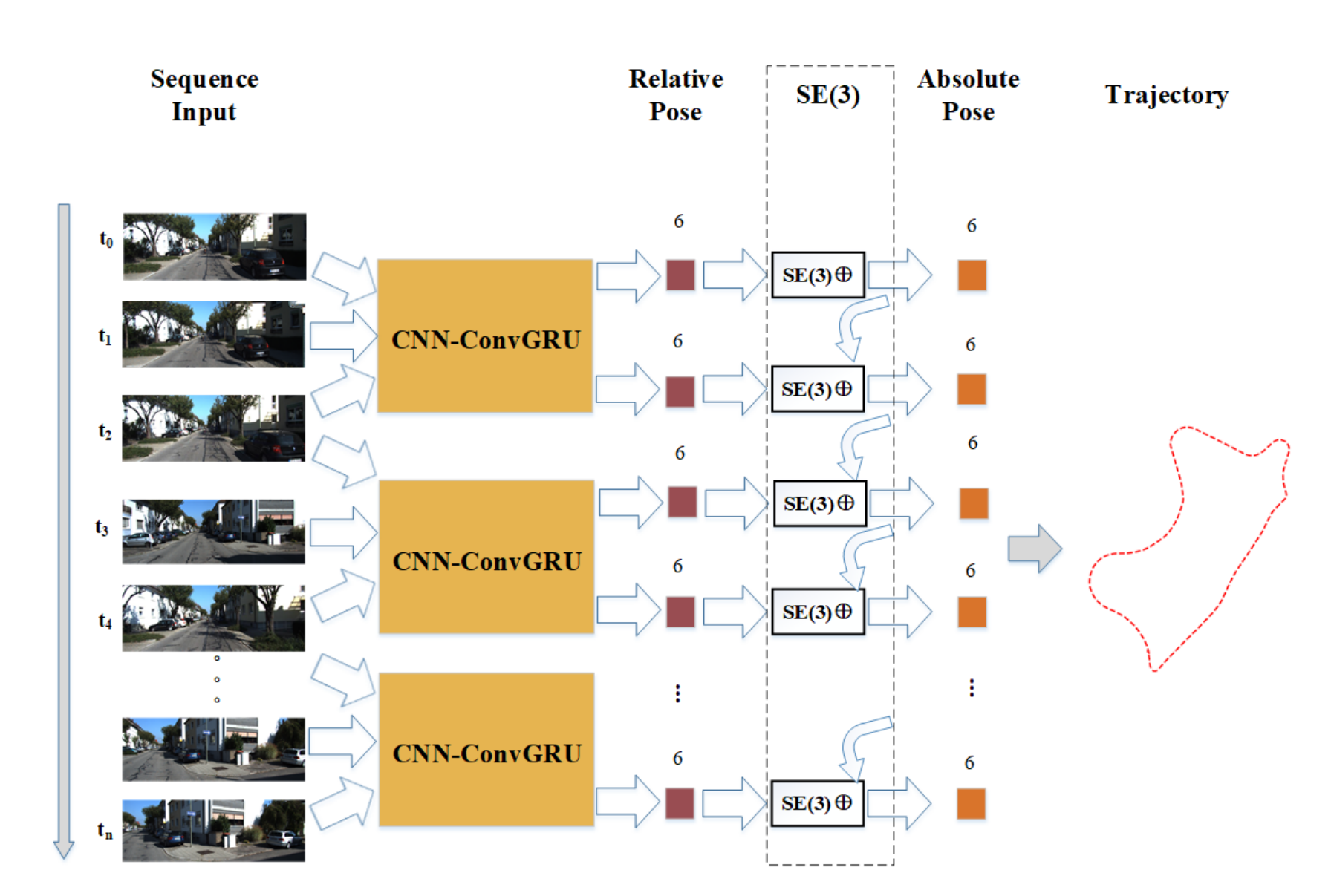}
    \caption{Our proposed end-to-end framework PoseConvGRU can estimate visual ego-motion by extracting geometrical feature among adjacent monocular RGB images. We can draw the trajectory after obtaining all the absolute poses.}
    \label{fig:PoseConvGRU}
\end{figure}

\section{Methodology}\label{Methodology}
The visual ego-motion problems are quite different from those of classification, tracking:

\smallskip
\noindent
\textbf{Firstly}, visual ego-motion estimation based on deep learning is a regression problem. It is not possible to accurately obtain the relative pose of two adjacent frames by simply identifying or detecting the objects in the images;

\smallskip
\noindent
\textbf{Secondly}, visual odometry problem needs to process two images at the same time, and it is especially related to the order of the images, because the relative poses between the two images can be numerically reciprocal from each other based on their respective references, so that we can obtain two various results.

Therefore, we can not simply use the popular neural network frameworks such as AlexNet \cite{krizhevsky2012imagenet}, VGGNet \cite{simonyan2014very}, GoogLeNet \cite{szegedy2015going}, ResNet \cite{he2016deep}, DenseNet \cite{iandola2014densenet} to solve this estimation problem, but should adopt the structure that can learn the geometric features of the images. The overall framework is shown in Fig.~\ref{fig:PoseConvGRU}.

\renewcommand\arraystretch{0.8}
\begin{table}[h]
\centering
\vspace{0mm}
\caption{CNN parameters. We can see the size of kernels decreases more as the depth of the network going deeper and the size of the feature maps decreases further.}\label{tab:cnn-parameters}
\end{table}
    \begin{tabular}{c|c|c|c|c}
    \hline
    Layer & Kernel size & Stride & Weigths & Tensor size\\
    \hline
    Input & - & - & - & 1280$\times$384$\times$6\\
    Conv1 & 7$\times$7 & 2 & 6$\times$64 & 640$\times$192$\times$64\\
    Conv2 & 5$\times$5 & 2 & 64$\times$128 & 320$\times$96$\times$128\\
    Conv3 & 5$\times$5 & 2 & 128$\times$256 & 160$\times$48$\times$256\\
    Conv3\_1 & 3$\times$3 & 1 & 256$\times$256 & 160$\times$48$\times$256\\
    Conv4 & 3$\times$3 & 2 & 256$\times$512 & 80$\times$24$\times$512\\
    Conv4\_1 & 3$\times$3 & 1 & 512$\times$512 & 80$\times$24$\times$512\\
    Conv5 & 3$\times$3 & 2 & 512$\times$512 & 40$\times$12$\times$512\\
    Conv5\_1 & 3$\times$3 & 1 & 512$\times$512 & 40$\times$12$\times$512\\
    Conv6 & 3$\times$3 & 2 & 512$\times$1024 & 20$\times$6$\times$1024\\
    Conv6\_1 & 3$\times$3 & 1 & 1024$\times$1024 & 20$\times$6$\times$1024\\
    Max-pooling & 2$\times$2 & 2 & - & 10$\times$3$\times$1024\\
    \hline
    \end{tabular}
\vspace{0mm}
\subsection{The Structure of PoseConvGRU} \label{p:PoseConvGRU}

\smallskip
\noindent
\textbf{Feature-encoding module.} 
\smallskip
In order to use the effective CNN structure to automatically learn the geometric relationship from two adjacent images, our approach leverages the network structure proposed by Dosovitskiy \etal - FLowNetSimple \cite{dosovitskiy2015flownet} , which ignores the decoder part in the network, only focusing on the front convolution encoder. In DeepVO, the output feature maps of Conv6-1 are directly input into subsequent modules, which not only hugely increases network parameters and magnifies the storage of GPU, but also makes the training complexity of the network expanded, so we add a layer of Max-pooling behind the Conv6-1 layer to reduce dimensions of the feature maps.

\begin{figure*}[htb]
    \centering{
    \scalebox{1.05}{
    \includegraphics[width=4.5in]{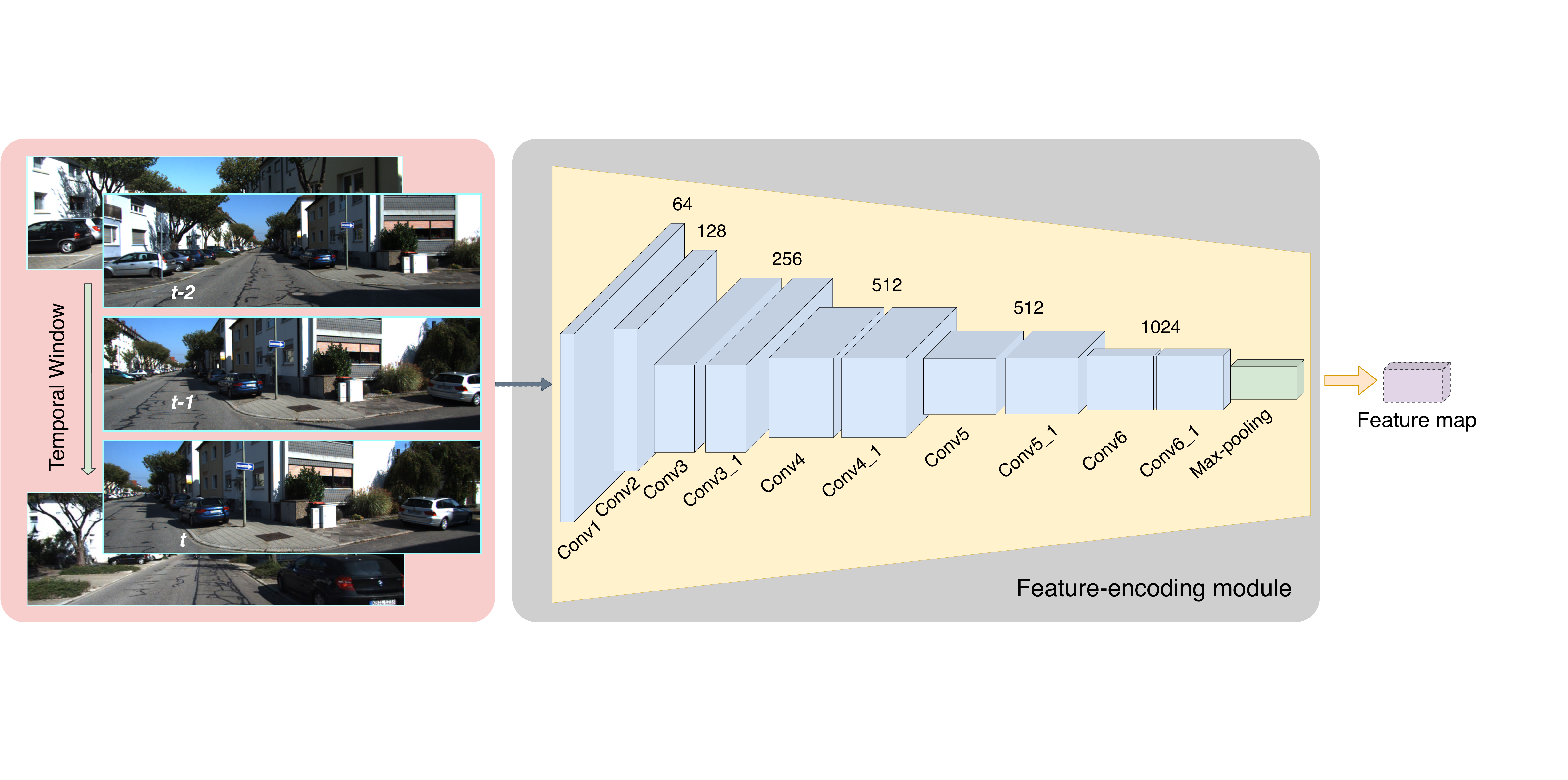}
    }}
    \caption{Feature-encoding module. We map RGB images temporally into this module to get output feature maps for estimating ego-motion further.}\label{fig:feature}
\end{figure*}
\vspace{-0mm}
The parameters of the CNN are shown in Table.~\ref{tab:cnn-parameters} . As shown in Fig.~\ref{fig:feature} , the convolutional neural network contains a total of 10 convolutional layers, and each layer is followed by a nonlinear activation function - ReLU (Rectified linear unit). The number of convolution kernels increases gradually as the depth of the network expands, so that more feature maps can be obtained, which can represent more abstract features, and the decreasing of the feature map size means that the CNN is paying more attention to large-scale and significant features. The size of the convolution kernel is also gradually reduced from 7 $\times$ 7 to 5 $\times$ 5 and finally to 3 $\times$ 3 for capturing local features. The input of CNN is the original continuous multi-frame RGB images, resized to 1280 $\times$ 384. Assuming that sequence’s length is n + 1, when adjacent two frames are combined in a sliding window, we can obtain n sets of image pairs. These image pairs are respectively subjected to 10 convolutional layers and the last Max-pooling layer to obtain feature maps of 10 $\times$ 3 $\times$ 1024 size. For multiple pairs of images generated by a time series, we refer to the structure of the Siamese network \cite{koch2015siamese} , using different branches to deal with similar problems, but will keep the CNN parameters weight-shared in the same time series, which means all the images of a sequence perform the feature extraction through the exact same CNN layer. We do not perform any pre-processing operations such as random clipping and rotating, to change the geometric relationship of the objects in the images, so that the original information of the images can be used for accurate pose estimation rightly.
 
\begin{figure*}[htb]
    \centering{
    \scalebox{1.05}{
    \includegraphics[width=4.5in]{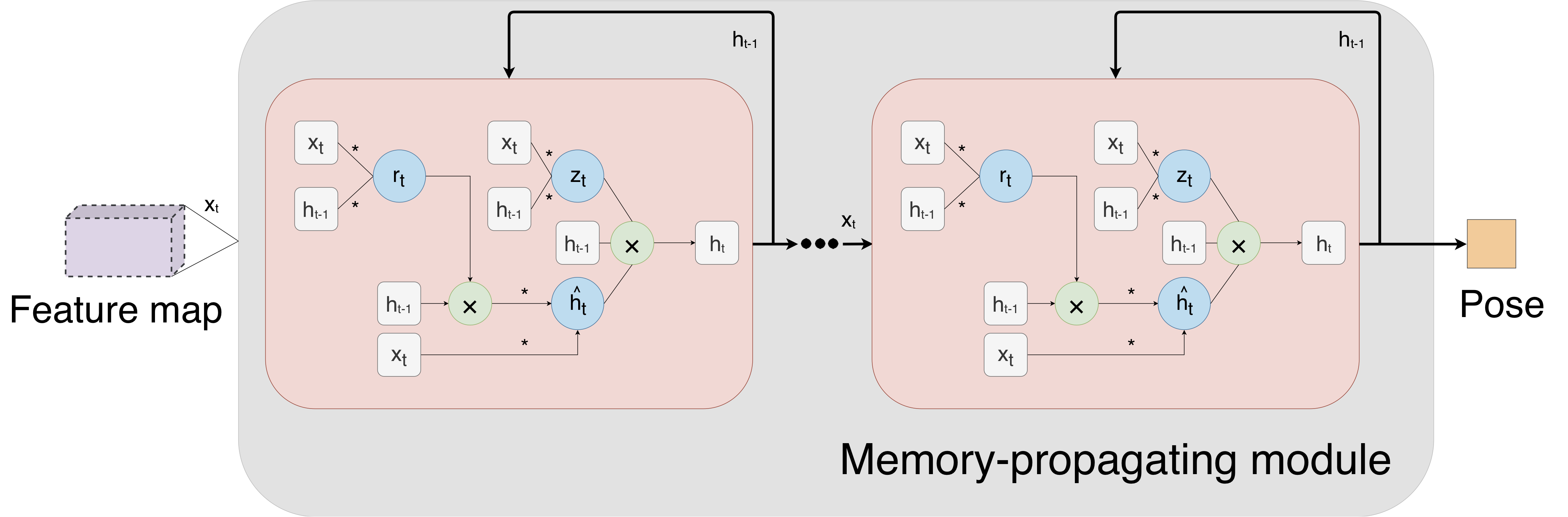}
    }}
    \caption{Memory-propagating module. We pass feature maps obtained from feature-encoding module into the stacked ConvGRU to propagate the long-term memory from video clips for camera poses regression.}\label{fig:memory}
\end{figure*}

\smallskip
\noindent
\textbf{Memory-propagating module.} 
The memory module builds a “visual memory” in the video clip, i.e., a long-term joint visual representation of all the clip frames to generate the transformation pose of each pair since it allows the neural network to automatically learn the intrinsic relationship among successive poses, module structure shown in Fig.~\ref{fig:memory}. We use a stacked ConvGRU (convolutional gated units) \cite{ballas2015delving} as our memory-propagating module, mathematical expression shown in the Equation.~\ref{convgru} \cite{siam2017convolutional}. On the one hand, ConvGRU can remember the states of historical moments, such as the geometric relationship coming from the previous frames of images, and then estimate the pose of the current moment utilizing the geometric constraint within multiple frames; on the other hand, we choose ConvGRU rather than ConvLSTM as our memory module because it is shown that GRU has similar performance to LSTM but with reduced number of gates thus fewer parameters \cite{chung2014empirical} . The image sequence is extracted by our feature-encoding module to obtain multiple 10 $\times$ 3 $\times$ 1024 tensors propagated into the stacked ConvGRU. In order to further improve the presentation capability and dynamic characteristics of the whole framework, 3 cells of ConvGRU are used in the practice. The end output, used for pose regression, will be a 6-dimensional pose vector, which respectively represents the relative poses $(\Delta x, \Delta y, \Delta z, \Delta \psi, \Delta \chi, \Delta \phi)$ between adjacent two images. Finally, we transform obtained pose vectors to SE(3) and calculate the absolute ego-motion.

\begin{equation}
\begin{aligned} z_{t} &=\sigma\left(W_{h z} * h_{t-1}+W_{x z} * x_{t}+b_{z}\right) \\ r_{t} &=\sigma\left(W_{h r} * h_{t-1}+W_{x r} * x_{t}+b_{r}\right) \\ \hat{h}_{t} &=\Phi\left(W_{h} *\left(r_{t} \odot h_{t-1}\right)+W_{x} * x_{t}+b\right) \\ h_{t} &=\left(1-z_{t}\right) \odot h_{t-1}+z \odot \hat{h}_{t} \end{aligned}
\end{equation}\label{convgru}

\subsection{Loss Function for PoseConvGRU}\label{Loss Functions}  

Visual ego-motion estimation problem can be regarded as a conditional probability problem: given an image sequence $X$ = ($X_{1}$, $X_{2}$, ..., $X_{n+1}$), calculated the appearance probability of the poses $Y$ = ($Y_1$, $Y_2$, ..., $Y_n$) between two adjacent images in this series.
\begin{equation}
p(Y | X)=p\left(Y_{1}, Y_{2}, \ldots, Y_{n} | X_{1}, X_{2}, \ldots, X_{n+1}\right)
\end{equation}
The problem to be solved here is how to decide the optimal network parameters $w^{*}$ to maximize the above probability.
\begin{equation}
w^{*}=\underset{w}{\operatorname{argmax}} p(X | Y ; w)
\end{equation}
So for M sequences, MSE (Mean Squared Error) is used as the error evaluation function, and the loss function that needs to be optimized finally can be obtained as
\begin{equation}
w^{*}=\underset{w}{\operatorname{argmax}} \frac{1}{M N} \sum_{i=1}^{M} \sum_{j=1}^{N}\left\|P_{i j}-\hat{P}_{i j}\right\|_{2}^{2}+\beta\left\|\Phi_{i j}-\hat{\Phi}_{i j}\right\|_{2}^{2}
\end{equation}
($\hat{P}_{i j}$, $\hat{\Phi}_{i j}$) represents the position and orientation of the image at the $j^{th}$ moment in the $i^{th}$ sequence relative to the image at the next moment in the sequence while $\beta$ is a scale factor used to maintain the balance between the position error and the orientation error. $\|\cdot\|_{2}$ represents a two norm. 

\subsection{Mirror-like Constraints through Data Augmentation}\label{Data Augmentation}
 
We further add some constraints into network by processing data augmentation along the training step aiming to perform better in tests.  

\smallskip
\noindent
\textbf{Data preparation.}
We take the sequences 00, 01, 02, 08, 09 in KITTI dataset \cite{Geiger2012CVPR} for training, which can not satisfy realistic requirements because of high-velocity situations, velocity-variable situations, or even car-reversing situations, so it is necessary to augment data for facing challenges above. Our data augmentation is performed on the fly. We augment the training data by randomly skipping frames to simulate the first and second challenge, which results in a better performance. Randomly horizontal flipping and temporal flipping of the sequences is also performed to allevate influnences caused by the last challenge.

\begin{figure*}[htb]
    \centering{
    \scalebox{1.05}{
    \includegraphics[width=4.5in]{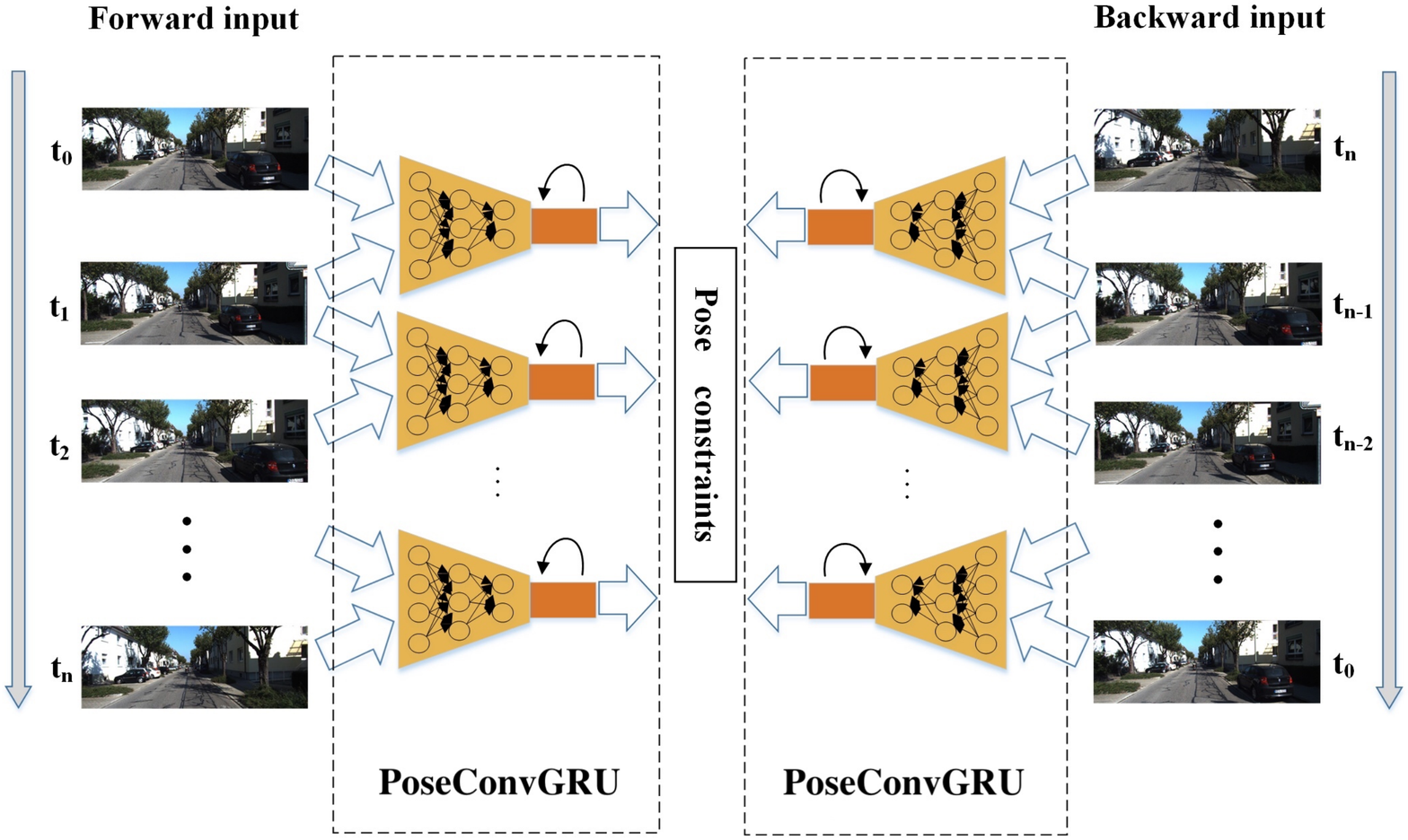}
    }}
    \caption{Constraints in practice. The structure of whole framework presents a mirror-like symmetric construction. All CNNs in this model are weight-shared.}\label{fig:constraints}
\end{figure*}

\smallskip
\noindent
\textbf{Advanced mirror-like constraints.}
PoseConvGRU-cons increases the accuracy of camera relative ego-motion estimation by adding some advanced constraints. Framework and specific implementation can be shown in Fig.~\ref{fig:constraints}.

The left half is exactly the same as PoseConvGRU. Image sequences pass through the feature-encoding module, Memory-propagating module and obtained outputs represent the relative poses between two adjacent frames. The right half and the left half are completely symmetrical, except that the image sequences RE input in reverse order, which are augmented in the data preparation process and the obtained outputs represent the relative poses of the previous frames relative to the current frames. The loss function based on MSE of all output poses, express as
\begin{equation}
\begin{aligned}
\textit {Loss}=\frac{1}{M N} \sum_{i=1}^{M} \sum_{j=1}^{N}&\left\|P_{1 i j}-\hat{P}_{1 i j}\right\|_{2}^{2}+\beta_{1}\left\|\Phi_{1 i j}-\hat{\Phi}_{1 i j}\right\|_{2}^{2}+\\
&\left\|P_{2 i j}-\hat{P}_{2 i j}\right\|_{2}^{2}+\beta_{2}\left\|\Phi_{2 i j}-\hat{\Phi}_{2 i j}\right\|_{2}^{2}
\end{aligned}
\end{equation}
($\hat{P}_{1 i j}$, $\hat{\Phi}_{1 i j}$) represents the position and orientation of the image at the $j^{th}$ moment of the forward input in the $i^{th}$ sequence relative to the image at the next moment in the sequence while ($\hat{P}_{2 i j}$, $\hat{\Phi}_{2 i j}$) represents the position and orientation of the image at the $j^{th}$ moment of the backward input in the $i^{th}$ sequence relative to the image at the previous moment in the original sequence. $\beta_1$ and $\beta_2$ separately represent the scale factors of the position error and the orientation error in the positive sequence input and the reverse sequence input. $\|\cdot\|_{2}$ represents l2-loss we use. 

\section{Experiment}\label{Experiment} 

In this section, we validate our proposed framework on the KITTI Visual Odometry / SLAM Evaluation dataset \cite{Geiger2012CVPR} and we take monocular VISO2-M and stereo VISO2-S \cite{geiger2011stereoscan} as our compared methods. Only the first 11 series of the KITTI dataset have the ground truth data of the images (sequence 00-10), so quantitative experiments can be performed in these 11 scenarios to compare the advantages and disadvantages of the various methods, noted that ablation experiments are proceeded along this part. 

The specific procedure to compare approaches is to firstly select the training sets, the validation sets and the test sets from these 11 scenarios that do not repeat each other. Then train models using the training sets along settling down hyper-parameters with the validation sets, and test models on the test sets. Finally we use the ground truth already collected to evaluate the error.

\subsection{Traning and Evaluating Protocols on PoseConvGRU}
We use the same method as Wang \etal used in the paper \cite{mohanty2016deepvo}, sequence 00, 01, 02, 08, 09 are used as the training sets; the remaining 6 scenes (sequence 03, 04, 05, 06, 07, 10) are used as evaluation sets. The validation sets is randomly selected from the training sets, following the principle of sampling without replacement. The specific data is shown in Table.~\ref{tab:train-test}.  A key issue here is how to generate sample sequences of images. In the experiment, we randomly select a frame of images as the starting frame, and then successively take several frames to form an image sequence of length $T_1$. In order to simplify the data training process, a fixed-length sequence is used, and the starting frames of two adjacent sample sequences are also selected across multiple frames, thereby avoiding excessive data overlap between samples. If we want to augment data set, we can sample sequences every other frame. One key rule of sampling is to ensure that there are enough identical scenes between the two images. If camera moves too fast in some frames, data augmentation must be handled carefully.

\renewcommand\arraystretch{1.5}
\begin{table}[htb]
\begin{center}
\caption{The components of dataset}\label{tab:train-test}
    \begin{tabular}{c|c|c}
	\hline 
    Dataset & Sequence & Total image pairs \\
    \hline
	Train & \multirow{2}{*}{00 01 02 08 09} & 15320 \\
	\cline{1-1} \cline{3-3}
	Validation & & 640 \\
    \hline
	Test & 03 04 05 06 07 10 & 7230 \\
	\hline
    \end{tabular}
\end{center}
\end{table}
\vspace{-5mm}
\smallskip
\noindent
\textbf{Training details.}
The entire network was built on the popular deep learning framework PyTorch-0.4.1. We trained our model using the AMSGrad \cite{sashank2018convergence} variant of Adam \cite{kingma2014adam}.The initial learning rate is $10^{-4}$ . As the number of training increases, the learning rate will be appropriately reduced to ensure that the optimization function is closer to the optimal solution. It uses two NVIDIA TITAN X (Pascal) GPUs for acceleration. The batch size is set to 32. It takes about 0.15 hours to train an epoch (all training data is trained for one time). After the end of an epoch, all training samples will be disordered to ensure that the training loss curve will drop smoothly. It takes about 20 hours for the entire experiment to achieve a small loss error. 

\begin{figure}[htb]
    \centering
    \includegraphics[width=4.5in]{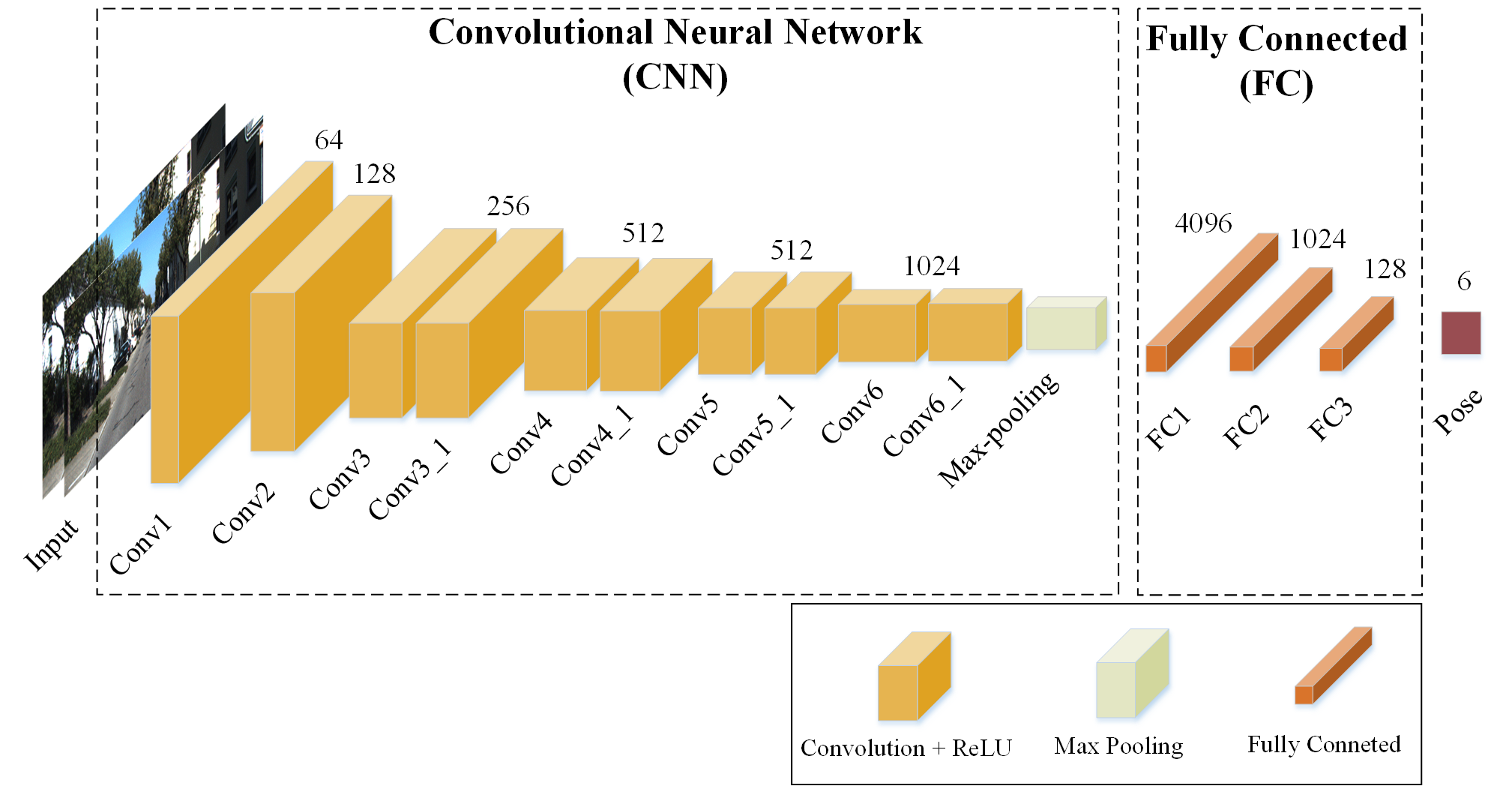}
    \caption{The structure of Ours-onlyNN. We leverage the FlowNetSimple structure, adding a extra max-pooling layer to estimate poses directly. Details can be seen in \ref{Ablation} .}
    \label{fig:CNN}
\end{figure}

In general, training the network combining these two modules is difficult to converge, in order to shorten the convergence time, feature-extracting module is pre-trained firstly. As shown in our ablation experiments part \ref{Ablation}, we actually only train CNN and FC layers to estimate camera poses (Recurrent Neural Network excluded), structure shown in Fig.~\ref{fig:CNN} . The Xavier Initialization method is used for setting weights in network parameters during training \cite{glorot2010understanding}, and the bias is zero-initialized. 

After pre-training process, we directly uses the parameters of the trained model in CNN as the initial parameters in the feature-extracting module of PoseConvGRU, that is, the fine-tuning operation.

Suppose the image sequence length is $T_1 = 11$ , and the ConvGRU's batch size is set to 4, so the batch size of the first module is $(11 - 1) \times 4 = 40$ . For each batch size, we sets the initial state of the GRU to 0. This is because, on the one hand, the image sequences we use are not selected according to the principle of no resampling. So without disordering the samples, the adjacent two batches will contain the same image frames, if the state here needs to pass through the two batch sizes, we need to ensure that the last frame of the first batch is the same as the first frame of the second batch, which however limits the diversity of the network’s samples. On the other hand, this sampling method is equivalent to dividing a scene into a number of irrelevant tracking  segmented series (including repeated images). So each sample can be learned the regulations separately. Setting the initial state of ConvGRU to 0 means that each sample was learned from the same state, and actually the experiments proved it is indeed feasible. There are also related methods to learn the initial state as a network parameter when training Recurrent Neural Network, but this is not necessary for our case.

\smallskip
\noindent
\textbf{Additional constraints illustrations.}
For PoseConvGRU with advanced constraints (PoseConvGRU-cons) , it is similar to the origin. The only difference is that the image sequences need to be input in positive ones and reverse ones to the network, and the relative poses between two adjacent images contain both the positive input and reversed input. When $T_1 , T_2$ are kept constant, the batch size in PoseConvGRU-cons is half of that of original PoseConvGRU. Other parameters remain unchanged during training. It takes about 0.5 hours to train an epoch. The entire experiment was trained for about 50 hours. 

\smallskip
\noindent
\textbf{Evaluation metrics.}
We evaluate our approaches on \textit{translation / rotation errors for subsequences} and \textit{translation / rotation errors for different speeds}, which are most commonly used evaluation metrics on the KITTI VO/SLAM dataset. The subsequences generally take 8 kind of lengths: 100m, 200m, ..., 800m. When calculating the error, we should sample the sequence of the same length from the entire trajectory, calculate the change of the camera poses on this sequence, and compare them with the ground truth, deriving translation and rotation error respectively, computing the average error of all the sample sequences as the average error of the current sequence. Finally we traverse the subsequence of 8 different lengths to obtain the average error of various subsequences.Another evaluation metric relies on the speed of the mobile platform. We take an average of 7 values from the lowest speed to the highest speed, calculating the error between the estimated poses and ground truth at different speeds.

\subsection{Ablation and Comparison Experiment}\label{Ablation}

In this section, we compare our proposed PoseConvGRU with monocular geometrical methods VISO2-M and stereo geometrical method VISO2-S and design ablation experiments along with other branches to verify each module's effectiveness as well as conduct several comparison experiments to reveal and verify implementation details in addition. All the following experiments are carried out on KITTI VO dataset.

\begin{figure}[htb]
    \centering
    \includegraphics[width=4.5in]{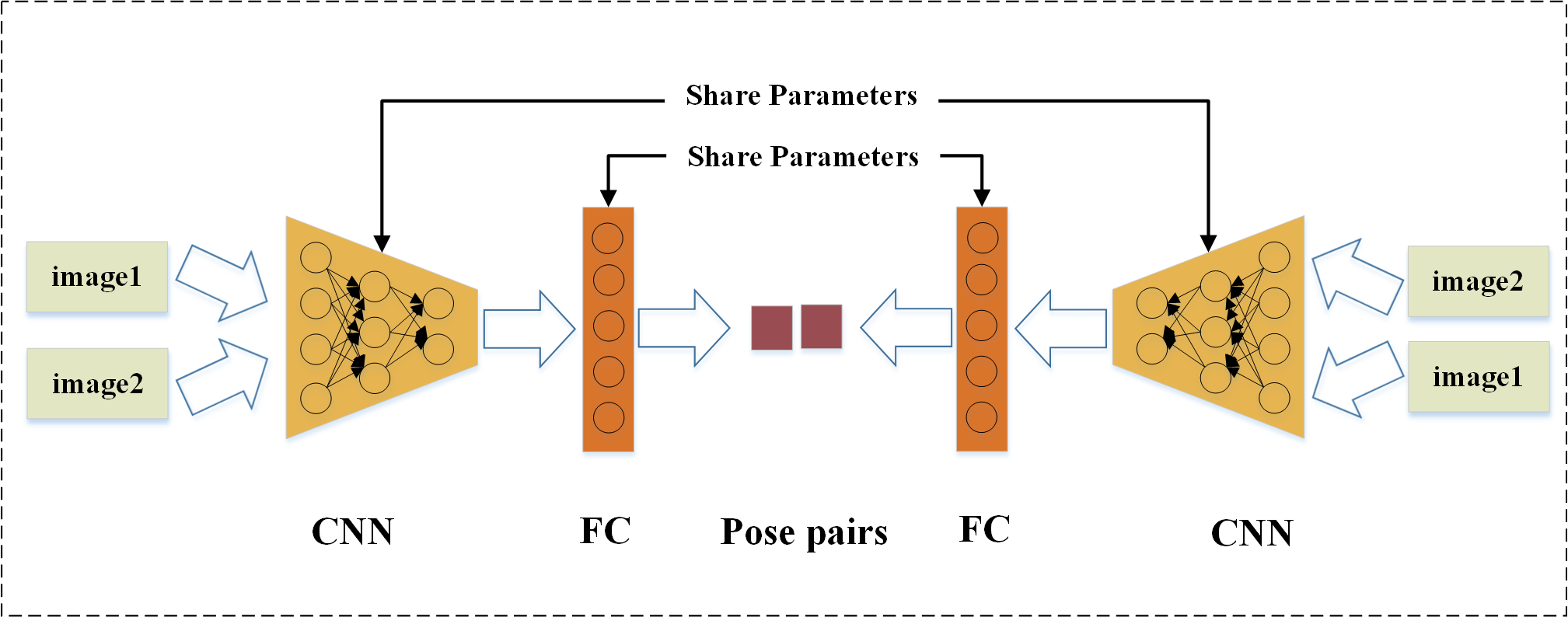}
    \caption{The structure of Ours-onlyCNN-cons. We do the same constraints as done in \ref{Data Augmentation} part to help proceed ablation experiment.}
    \label{fig:CNN-cons}
\end{figure}

We actually firstly pre-train CNN and some FC layers adding constraints to estimate visual ego-motion directly, constrained structure performed in Fig.~\ref{fig:CNN-cons} then we remove FC layers remaining CNN along with parameters as our initial feature-encoding module in PoseConvGRU. Next we proceed fine-tuning with training stacked ConvGRU to achieve our whole framework. As a result, we derive four branches named Ours-onlyCNN, Ours-onlyCNN-cons, Ours, Ours-cons.

\begin{figure*}[htb]
    \centering{
    \scalebox{1.05}{
    \includegraphics[width=4.5in]{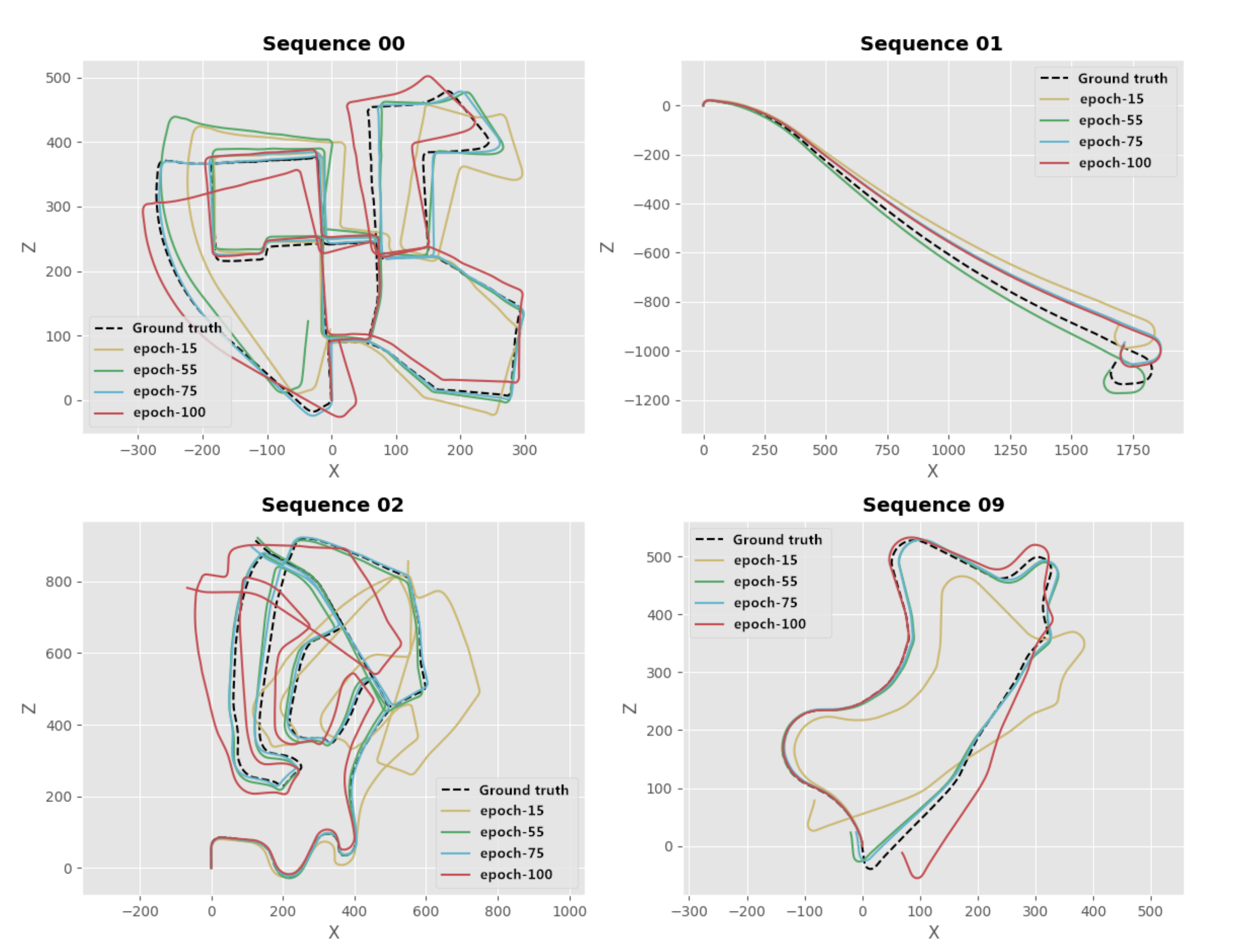}
    }}
    \caption{Trajectories under different \textit{epoch} (training sets) (Best viewed with zoom-in.)}
    \label{fig:epoch_train}
\end{figure*}

\begin{figure*}[t]
    \centering{
    \scalebox{1.0}{
    \includegraphics[width=4.5in]{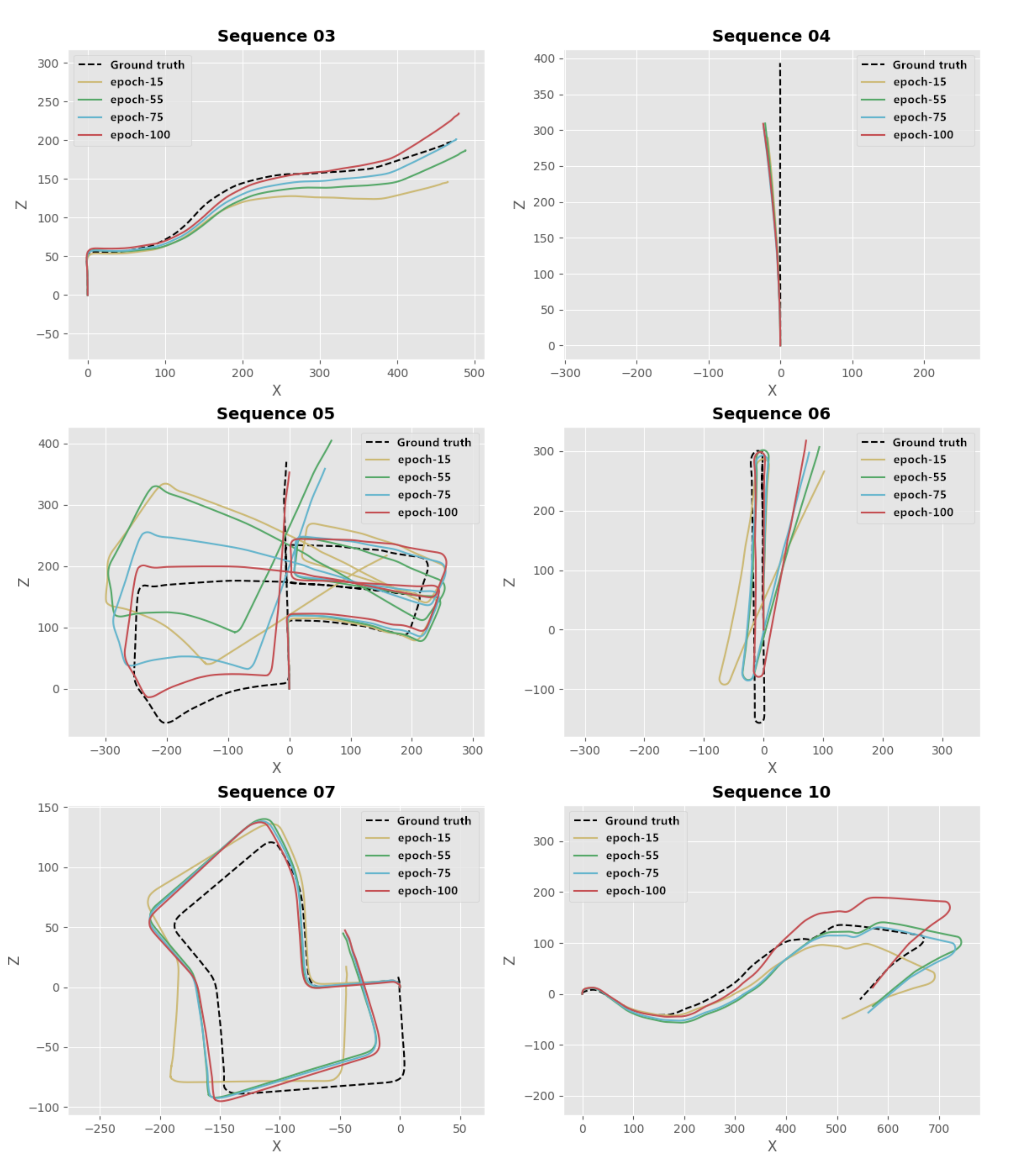}
    }}
    \caption{Trajectories under different \textit{epoch} (test sets) (Best viewed with zoom-in.)}
    \label{fig:epoch_test}
\end{figure*}

\begin{figure*}[htb]
    \centering{
    \scalebox{1.05}{
    \includegraphics[width=4.5in]{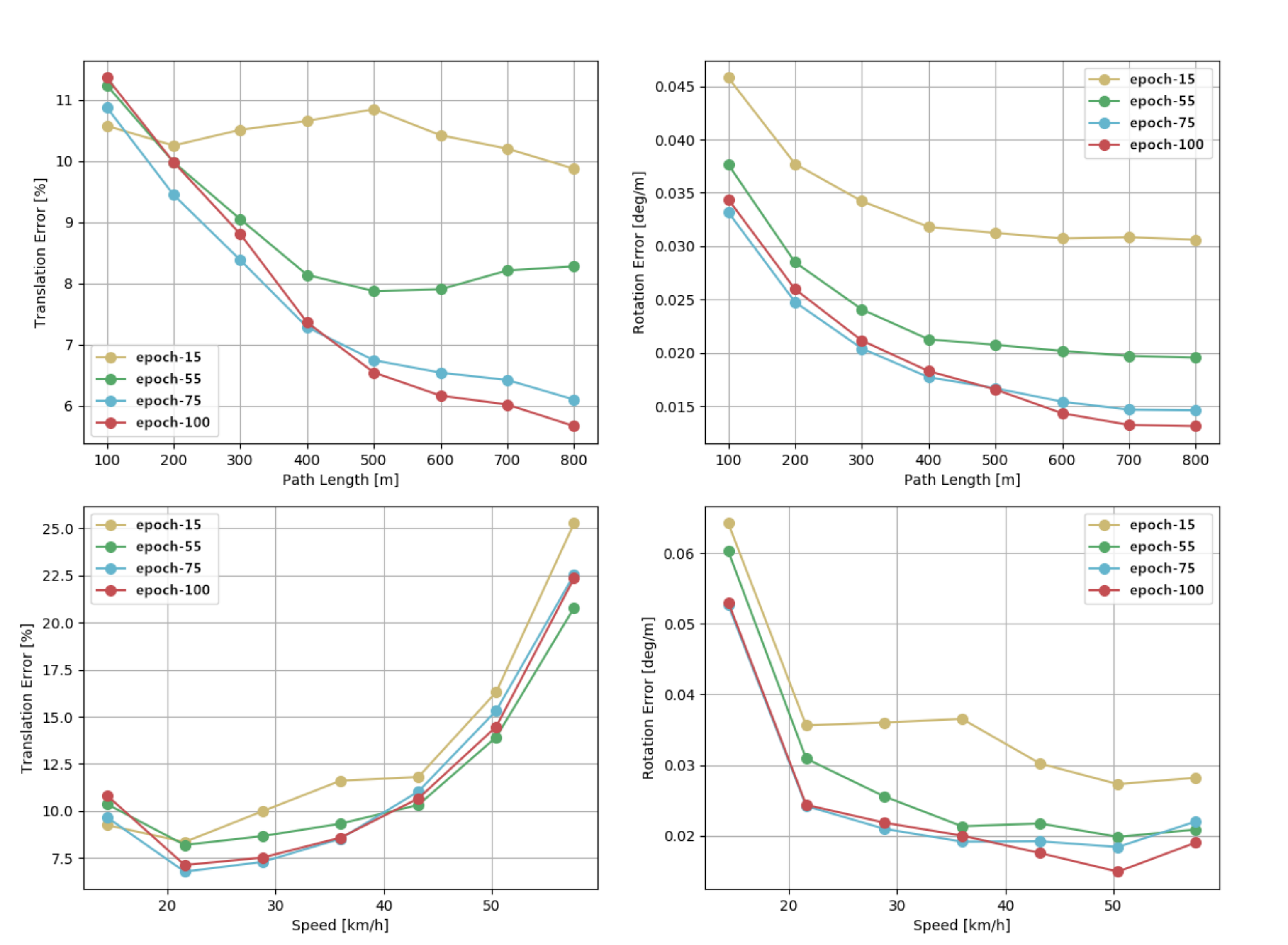}
    }}
    \caption{Comparison results (test sets) under different \textit{epoch}}
    \label{fig:epoch_test_2}
\end{figure*}

\smallskip
\noindent
\textbf{Effectiveness on \textit{epoch} in pre-training process.} Because the pre-training process is quite long, in order to test the model in the middle of training, the model parameters are saved every few epoch. Taking Ours-onlyCNN-cons as an example, the whole training process last for about 100 epochs, and the models with epoch=15, 55, 75, 100 were taken out for testing. The trajectory curves on the training set are shown in Fig.~\ref{fig:epoch_train}. The trajectory curves on the test set are shown in Fig.~\ref{fig:epoch_test}, and the curves of the evaluation metrics on the test set are shown in Fig.~\ref{fig:epoch_test_2}.

Fig.~\ref{fig:epoch_train} shows that at the initial stage of training, since the translation and rotation error are both large, the trajectory of epoch-15, with the error between frames gradually accumulating, gets more and more away from the real one. As the number of \textit{epoch} increases, the trajectory on the training set and the ground truth get closer and closer, and the epoch-75 almost perfectly coincides on the sequence 00, 02, 09, which is also consistent with the continuous decline in training error. When the number of iterations is further increased, it is found that the trajectory of epoch-100 on sequence 02, 08 deviates far from the counterparts. In fact, the training error at this moment is lower than that of the previous ones. It happens because the loss function describes only the MSE of the poses between two adjacent frames. When the poses of some frames deviates far away from the ground truth, even if the poses of other frames is estimated accurately, the trajectory still drift far away. If we continue to train, the training error will be further reduced, and the scene trajectory will surely fluctuate around the ground truth, but the variation will be more modest. However, due to limited time and the convergence performance for the training scene does not represent the generalization ability of the model, it is also necessary to consider the presentation on the test sets.

Fig.~\ref{fig:epoch_test} shows the trajectory curves for different sequences in the test sets. It can be clearly seen from sequence 03, 05 that as the number of \textit{epoch} increases, the estimated trajectory becomes closer to the ground truth, indicating that the framework designed in this paper can learn the motion relationship between two adjacent frames. Even if many scenes are not seen before, the camera ego-motion can also be estimated well, and do not have overfitting problem. An important reason is that we uses tricks such as dropout when training the network. For sequence 04, the model test results under different epoch times are almost the same. It can be seen that the camera basically moves straight forward, and the rotation change is modest, so the translation estimation is very sensitive to this scene. It will be better to design network for the translation and rotation separately. For sequence 06, 07, The test results of the models saved by epoch-55, epoch-75 and epoch-100 basically have no obvious performance improvement, illustrating that the network’s learning ability has reached the bottleneck, and even if the number of \textit{epoch} is increased, the performance will not increase significantly. For sequence10, the estimated trajectory performs better on epoch-55 and epoch-75 than epoch-100, indicating that the model may have overfitting problem for this scenario. In order to further analyze the performance of the model, it is also necessary to compare the quantitative test metrics between the models.

The four graphs in Fig.~\ref{fig:epoch_test} represent the former error evaluation metrics on the test sets: the upper left graph shows the translation error subsequences under different length, the upper right graph shows the rotation error under different length subsequences, and the lower left graph shows the translation errors at different speeds. The lower right graph shows the rotation error at different speeds. It can be seen that for subsequences of the same length, as the number of training \textit{epoch} increases, the translation error and the rotation error are significantly improved, but there is basically no large improvement after epoch-75. For the ego-motion estimation at the same speed, the translation error of the ultimate training (epoch-75) is less obvious than it of the initial training (epoch-15), especially when the speed exceeds 40 $km/h$ , the effect of epohc-75 is worse than that of epoch-55. One big reason is that the speed distribution in the scenes of the training sets and the test sets is different. Therefore, if there is a lack of image pairs with higher speed in the training sets, test sets will not accurately estimate the high-speed translation. To alleviate this problem, we can try to expand the sample (sample in every few frames) in the training sets, thus improving the generalization ability of the model.

\smallskip
\noindent
\textbf{Quantitative Experiment result.} For PoseConvGRU, set $\beta$ = 0.9, $T_1$ = 3, $T_2$ = 2, network parameters iterated 90 epochs were used for testing. For PoseConvGRU-cons, set $\beta_{1}=0.9$, $\beta_{2}=0.999$, network parameters iterated 80 epochs were used for testing. The trajectory on the test set is shown in Figure 4.14, and the evaluation metrics on the test set are shown in Figure 4.15.

\begin{figure*}[htb]
    \centering{
    \scalebox{0.95}{
    \includegraphics[width=4.5in]{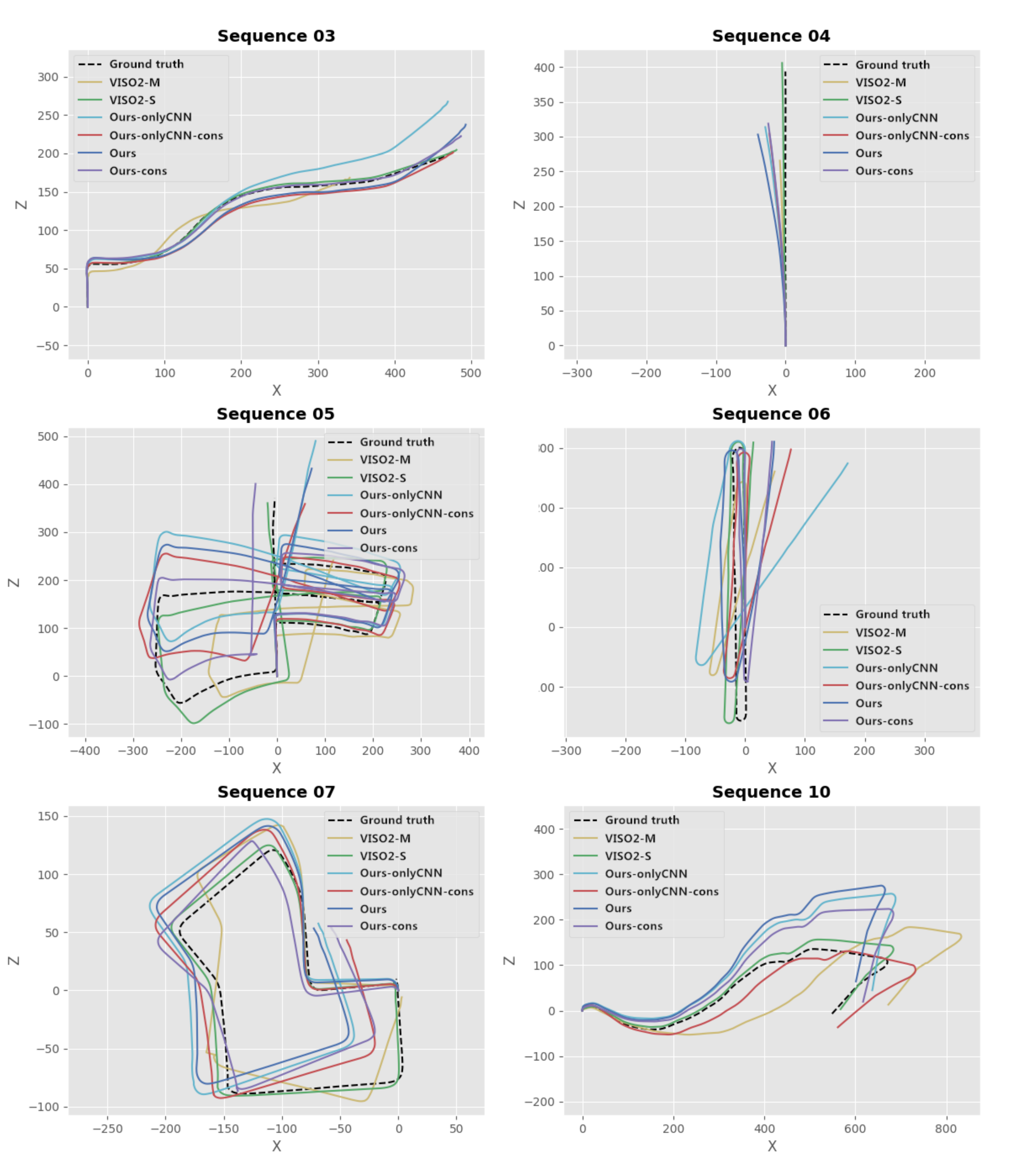}
    }}
    \caption{Trajectories of our methods and LIBVISO2 on the Sequence 03, 04, 05, 06, 07, 10 (test sets) (Best viewed with zoom-in.)}\label{fig:test-1}
\end{figure*}

The experiment totally compared six methods, including two baseline methods VISO2-M and VISO2-S, Ours-onlyCNN and Ours-onlyCNN-cons, Ours and Ours-cons. As can be seen from Fig.~\ref{fig:test-1}, the six methods can slightly estimate the shape of the trajectory accurately, but the details are various from each other, and it is difficult to compare the advantages and disadvantages of various methods from a certain situation. For example, PoseConvGRU-cons fits the real trajectory better on the Sequence 03 than the onlyCNN, but is slightly inferior to the onlyCNN-cons on the Sequence 05.

\begin{figure*}[htb]
    \centering{
    \scalebox{1.05}{
    \includegraphics[width=4.5in]{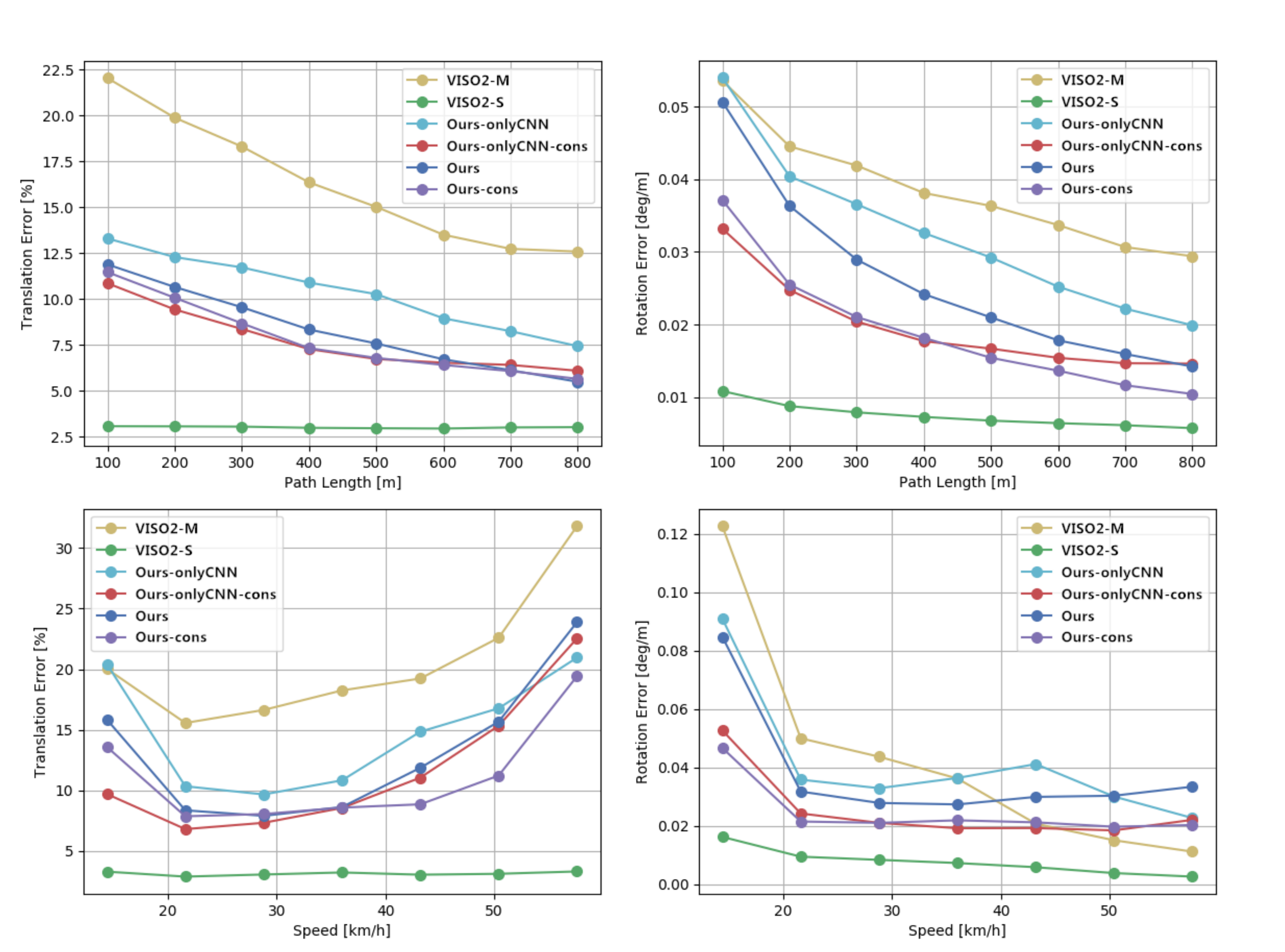}
    }}
    \caption{Comparison results of different methods on the Sequence 03, 04, 05, 06, 07, 10 (test sets)}
    \label{fig:test}
\end{figure*}

Fig.~\ref{fig:test} shows the error of the six methods on different evaluation metrics, which can objectively reflect the performance of various methods. PoseConvGRU is basically better than onlyCNN on the evaluation of four kinds of errors, because PoseConvGRU is fine-tuning on the basis of onlyCNN, and joins ConvGRU for multi-frame constraints. ConvGRU does improve the accuracy of visual ego-motion estimates. PoseConvGRU-cons is superior to PoseConvGRU in these evaluation metrics, indicating that the performance of the network can be improved by adding these advanced constraints. However, the effect of PoseConvGRU-cons is comparable to that of onlyCNN-cons. The main reason we believe is that the sample size is far from enough, because both PoseConvGRU-cons and onlyCNN-cons is to double the training data. Yet, the training data in our paper is limited, so we can only obtain different samples by combining the existing images. This does not essentially introduce new sample images, so even though we constrain our framework with multi-frame through ConvGRU, it is impossible to improve the generalization ability of the network further.
The performance of the four deep learning methods proposed in this paper is better than the monocular VISO2-M, while the latter two methods with constraints are slightly inferior to the stereo VISO2-S. Our future work consider taking into account the stereo RGB image to achieve better effect.

\smallskip
\noindent
\textbf{Qualitative Experiment result.}
Quantitative experiments only use the first 11 sequences with ground truth to proceed training and testing, which can quantitatively analyze error. Considering that the latter 11 scenes in KITTI do not provide ground truth, we can still use the trained model to test in these scenes and visually evaluate the effect of our method. In this paper, all the images of sequence 00-11 are used as the training sets, with a total of 23190 image pairs, taking a small part as the validation sets, and then we adopt the method in quantitative experiment to train it. When the model converges to a certain extent, the training sets and the validation sets are combined for fine-tuning until the training error is basically not reduced. The model at this time is taken as the final model of the qualitative experiment. Then use this model to test on sequence 11-21. We only show the performance of PoseConvGRU and onlyCNN.

It takes about 0.2 hour to train an epoch, and the whole experiment takes 20 hours to achieve a small training error. We saved the model parameters of the epoch-100 for subsequent testing. When training PoseConvGRU, the pre-processing of the dataset is consistent with that of onlyCNN: all scenarios of sequence 00-10 are regarded as training sets for our framework. In order to speed up the convergence time of the network, we uses the parameters of the pre-trained model in onlyCNN as the initial parameters of the feature-encoding module. The other parameter settings during training are exactly the same as the PoseConvGRU in quantitative experiment. An epoch takes about 0.4 hour, and the whole experiment takes 40 hours to achieve a small training error. 
\begin{figure*}[htb]
    \centering{
    \scalebox{0.95}{
    \includegraphics[width=4.5in]{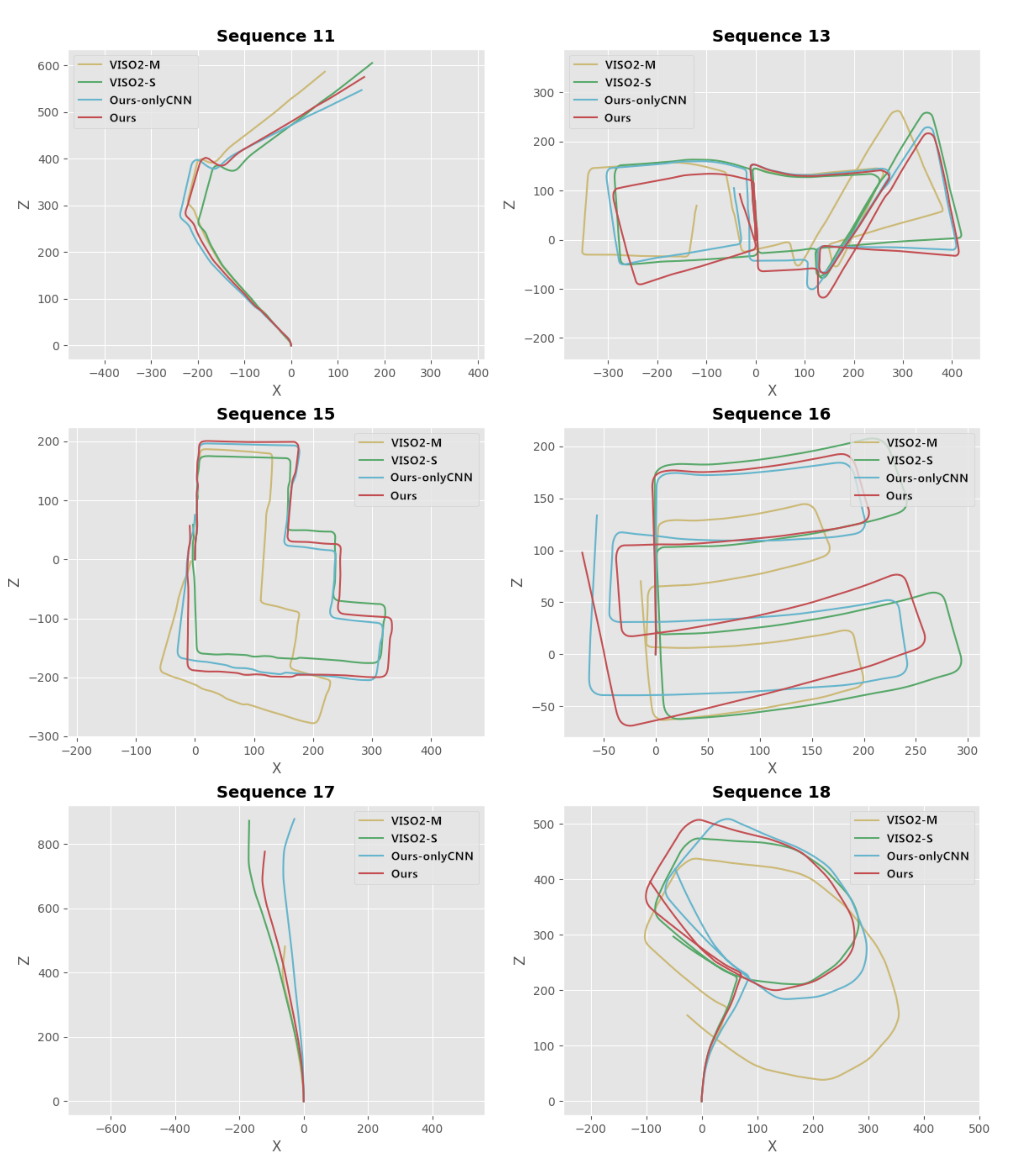}
    }}
    \caption{These six sequences do not have ground truth, so we proceed the qualitative experiments. (Best viewed with zoom-in.)}\label{fig:experiment2-test-1}
\end{figure*}
We tested the trained onlyCNN and PoseConvGRU models on the sequence 11-21. The partial results are shown in Fig.~\ref{fig:experiment2-test-1}, which also include the test results of two baseline methods VISO2-M and VISO2-S. From the scale of the scene, the range of these 11 scenes is from 100m × 100m (such as sequence 14) to 2000m × 5000m (such as sequence 21), and is quite different with training sets. From the perspective of vehicle speed, the top speed of the first 11 scenes is basically around 60km/h, while in the latter 11 scenes, the running speed of the vehicle is even as high as 90 $km/h$ . When the camera frame rate is fixed at 10 $fps$ , the movement between two adjacent images will become intense, and the lack of such samples in the training sets will affect the generalization ability of the deep learning.

Because of the lack of real trajectory data, we use the results of the VISO2-S method as a reference. In sequence 11, 13, 15, 16, 17, 18, the effect of CNN-VO-cons is better than that of CNN-VO and VISO2-M, and the trajectory of CNN-VO is also more accurate than VISO2-M. In most scenarios, the VISO2-M trajectory has large deviations in scale compared to the other three methods. This shows that the scale problem of monocular VISO2 in geometric method is a very big obstacle in pose estimation, and deep learning method can basically overcome this difficulty.

\section{Conclusion}\label{Conclusion}
We propose a novel data-driven Long-term Recurrent Convolutional Neural Networks (PoseConvGRU) encoding geometrical features in images to gauge camera poses, which is completely end-to-end. Our proposed neural network is more real-time and less calculation-consuming, unlike other learning-based ego-motion estimation algorithms, which need to spend plenty of time to calculate the pre-processed dense optical flow before training the neural network or use a pre-trained network to estimate the optical flow with additional calculation costs. The main idea is to use CNN to extract the geometric relationship features of two adjacent images in the image sequences along with data augmentation, then pass the feature maps through a stacked ConvGRU module for feature learning on the time series, and finally achieve the regression of the relative poses among consecutive multi-frame images. The performance of our approach is better than VISO2-M, a traditional geometric monocular method facing VO problem. In the future, we plan to focus on the stereo study of end-to-end visual ego-motion estimation, since some significant information like scale can be directly obtained from stereo images.

\section*{Acknowledgment}
This research was carried out at the Institute of Cyber-Systems
and Control, Zhejiang University, China. 

\section*{References}

\bibliography{dlvo}

\end{document}